\documentclass{easychair}
\usepackage{doc}

\usepackage{multicol}
\usepackage{xcolor}
\usepackage{framed}
\usepackage[utf8]{inputenc}
\usepackage{graphicx}

\usepackage{amsfonts}
\usepackage{amssymb}
\usepackage{amsmath}

\usepackage{enumitem} 
\usepackage{booktabs}

\newcommand{\commentOut}[1]{}

\newcommand{\bequ}{\begin{quote}}
\newcommand{\enqu}{\end{quote}}
\newcommand{\bece}{\begin{center}}
\newcommand{\ence}{\end{center}}

\newenvironment{compactitem}{\begin{itemize}}{\end{itemize}}

\title{Symbolic Informalization:\\ Fluent, Productive, Multilingual}
\titlerunning{Symbolic Informalization}

\author{Aarne Ranta}
\authorrunning{Aarne Ranta}

\institute{
  Department of Computer Science and Engineering,\\
  Chalmers University of Technology and University of Gothenburg,\\
  email aarne.ranta@cse.gu.se 
  }

\usepackage{booktabs} 
\usepackage{graphicx}
\usepackage[utf8]{inputenc}  

\begin{document}

\maketitle

\noindent
Symbolic informalization enables a reliable conversion of formal mathematics to natural language.
It has the potential to make machine-checked content human-readable without loss of precision.
In a traditional proof system usage, symbolic informalization generalizes the limited mechanisms of syntactic sugar into the ordinary language of mathematics.
In a setting where proofs are constructed by artificial intelligence and autoformalization, symbolic informalization can explain what precisely has been constructed.
This paper outlines the project Informath, which aims to show how symbolic informalization can produce fluent text with a reasonable development effort and address multiple formal and natural languages.
Informath is based on an interlingual architecture, where Dedukti works as a hub between different proof systems (Agda, Lean, Rocq) and Grammatical Framework (GF) takes care of linguistic correctness and variation in different natural languages.
%
%


\noindent
A slightly shorter version of this paper will appear in Roussanka Loukanova (ed.), \textit{Mathematical and Computational Linguistics for Proofs, MCLP 2025}, Studies in Computational Intelligence (SCI), Springer Nature, 2026.

\section{Introduction}

\label{introduction}

\textbf{Informalization} is the process of translating from formal to informal language --- in particular, from formal logic to \textbf{ordinary mathematical language}, by which we will mean the mixture of words and symbols used in textbooks and research papers.
Informalization is the inverse of \textbf{formalization} and often considered easier and less interesting.
What makes it easier is that formal languages are closed systems defined by explicit rules, whereas informal languages have no formal rules or clear-cut limits.

The following picture illustrates the situation.
It uses solid lines to mark precisely defined objects and procedures, dashed lines to mark vaguely defined ones, a solid arrowhead to mark a total function, and a hollow arrowhead to mark a partial function.

\begin{center}
\includegraphics[width=0.78\textwidth]{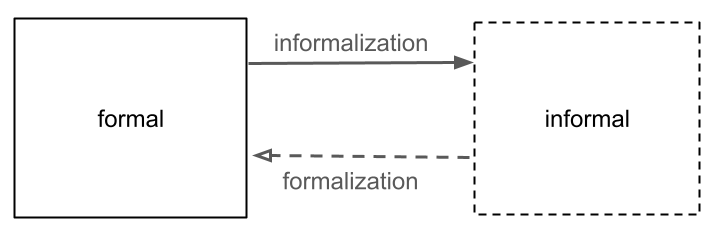}
\end{center}

\noindent
Informalization is in principle possible by \textbf{symbolic} methods, by a set of rules that covers the entire formal language and a well-defined part of the informal language. 
Advances in machine learning, in particular \textbf{large language models} (LLMs), suggest that formalization might be better approached by \textbf{neural} methods than by hopelessly incomplete symbolic rules.
Since the target is a formal language, the results of formalization can be automatically verified by a proof checker, which eliminates the problem of ``hallucinations'' typical of LLMs.
Thanks to this, recent successes in \textbf{autoformalization} have enabled the formalization and verification of entire mathematics books \cite{urban-2026}.

However, neural autoformalization cannot guarantee that the formalization corresponds to the original informal text.
For example, the theorem that is proved formally may be different from the theorem in the original text.
Since there is no formal definition of the informal language, there is no automatic way to compare the original statement and is formalization.
Instead, humans may need to compare the code with the text, which can be a daunting task.
Alternatively, the LLM can be used to informalize the code or parts of it --- but this is also a source of errors, because LLMs can ``hallucinate''.

Symbolic informalization comes in here as a natural complement to neural autoformalization.
It enables the human user to communicate with a formal system with the mediation of an LLM without the need to see the formal code.
The user can pose problems in ordinary language and get results in ordinary language; what happens in between is that an LLM formalizes the problem and builds a solution with a theorem prover, and the resulting proof is informalized back to her. 
The user can then check that the theorem is interpreted correctly and understand the constructed proof without having to look at the formalism.

The workflow just described sounds ideal: mathematicians can use proof systems in their ordinary language.
An LLM helps them to find proofs, a proof system checks them, and a symbolic informalization system --- which is a compiler-like component that fully covers the formalism --- explains the results to them.
Unfortunately, informalization is not as easy as one might expect.
In \cite{jiang-al-2024}, the authors state three reasons not to use symbolic informalization tools: these tools

\begin{compactitem}
\item ``result in natural language content that lacks the inherent diversity and flexibility in expression: they are rigid and not natural-language-like'',
\item ``are hard to design and implement'',
\item ``differ a lot for different formal languages, hence the approach is not scalable for multiple formal languages''.
\end{compactitem}
\noindent
All of these claims (except perhaps the third one) are generally believed in the natural language processing (NLP) community.
But they are also an excellent way to express our problem specification, as stated in the title of this paper.
They summarize three desirable properties of a symbolic informalization system: it must be

\begin{compactitem}
\item \textbf{fluent}: result in natural language content that shows the inherent diversity and flexibility in expression,
\item \textbf{productive}: feasible to design and implement,
\item \textbf{multilingual}: shared for different formal languages.
\end{compactitem}
In the following, we will address these three goals in a reverse order, from the easiest to the hardest.
The multilinguality goal has a relatively straightforward technical solution, whereas productivity needs some tool development.
Fluency is a genuine research problem, which definitely leads us to revise the expectation that informalization is ``easy''.

As our response to the challenge, we will introduce the system Informath \cite{informath-homepage}, which aims to provide a general resource that can be plugged in into different proof systems.
In that function, Informath can be seen as an ultimate implementation of syntactic sugar and user-defined notations.
In addition to explaining the results of autoformalization, Informath can help the usual interaction with proof systems. This works for both input and output: even though the primary purpose of Informath is informalization, it uses reversible techniques that can also be used in rule-based formalization.

The structure of the paper is as follows:
Section~\ref{background} gives historical background and related work.
Section~\ref{multilinguality} shows how to make informalization multilingual.
Section~\ref{productivity} introduces methods that enhance productivity.
Section~\ref{fluency} explains how the fluency problem is addressed.
Section~\ref{atour} describes the language currently covered by Informath.
Section~\ref{results} gives a summary of results.
Section~\ref{conclusions} states conclusions, with a focus on the three keywords of the title.
Section~\ref{futurework} lists some future work.

\section{Background}

\label{background}

Early attempts include the pairing of Automath \cite{debruijn-1980} with a Mathematical Vernacular \cite{debruijn-1994}, the Mizar system capable to produce journal articles \cite{bancerek-al-2015}, and the extraction of text from Coq (later Rocq) proofs \cite{coscoy-al-1995}.
Mathematical Vernacular was mostly a theoretical and pedagogical project, with no known implemented translation tools, whereas Mizar and the Coq system came with implementations.
Later systems with similar aims include Isabelle/Isar \cite{wenzel-1999}, GF-Alfa \cite{hallgren-ranta-2000} working on an early version of Agda, and Verbose Lean \cite{massot-2024}.
In all these systems, the output still tends to be hard to read when the complexity of the statements and proofs grows.
They are far from using all of the expressive power that natural language could provide to enhance readability.

In the direction of formalization, a prominent approach is \textbf{controlled natural languages} (CNL), which are fragments of natural languages defined by formal grammars.
Examples are Naproche \cite{cramer-al-2009}, ForTheL \cite{paskevich-2007}, and MathNat \cite{humayoun-raffalli-2010}.
Their formal grammars support a compiler-like syntax-directed translation from the CNL to a formal language.
Many such systems were originally designed as pedagogical tools for learning mathematics.
But Naproche, in particular, has later also been used in advanced mathematics development \cite{koepke-2025}.

Yet another approach is to deny the assumption that humans do not want to use formal notations.
In this alternative, one extends the formalism with informal math-like notations such as user-defined infix operators in Rocq \cite{bertot-casteran-2004}, generalized to mixfix in Agda \cite{norell-2007}, and to even more profound syntax extensions in Lean \cite{demoura-al-2015}.
However, this approach is still very limited compared to the ordinary language of mathematics.
It can also lead to dialects of formal languages that are hard to understand for outsiders, because of idiosyncratic notations different from standard mathematics.
What is more, extending the syntax of formal languages makes them more difficult to reason about and thereby open to errors.
The philosophy followed in Informath is to keep the formal system as simple as possible and separate the ``syntactic sugar'' into a proper informalization component. 

The first version of Informath was published in 2024 \cite{ranta-2024}.
It builds on earlier work using similar methods \cite{Ranta94A,ranta-1996,ranta-1997,hallgren-ranta-2000}, in particular, the architecture of \cite{ranta-2011c}.
However, the earlier work was mostly in the CNL tradition, where it is enough to produce \textit{some} linguistically correct expressions for any given formal content.
These approaches were challenged in \cite{ganesalingam-2013}, which declared the programme of building a formal analysis of everything that is expressed in the language ordinarily used by mathematicians.
The mission of Informath is to meet this challenge by constructing a grammar that gradually approaches the full language of mathematics.

\section{Multilinguality}

\label{multilinguality}

A scalable way to make a system multilingual is to make it \textbf{interlingual}.
This means that there is an intermediate language, an \textbf{interlingua}, in which the main part of processing (such as type checking) is performed.
The interlingua is typically formal and abstract, and not perceived directly by the users of the system.
But it is linked to the users' languages by mappings in both directions.
To add a new language to the system, one only needs to define these mappings.
In an ideal case, both directions are covered by a reversible, relational mapping.

Interlingua is an old idea in the translation of natural languages, dating back at least to Descartes, who suggested a universal language that would
\bequ
\textit{establish an order among all thoughts that can enter in the human spirit, in the same way as there is a natural order among numbers, and as one can learn in one day the names of all numbers up to infinity and write them in an unknown language, even though they are an infinity of different words.}
\enqu
\cite{descartes-1629}.
The first problem in these systems is how to define the interlingua itself: it must be able to express everything that one wants to express, with the same mathematical precision that exists among numbers.
Descartes might have been encouraged to hope for this by his success in bringing such precision to geometry.

Basically all of mathematics was later brought under this discipline in formal logic.
The most general device for this are \textbf{logical frameworks}, which can express theorems, proofs, and computations in single systems \cite{harper-al-1993}.
A system particularly designed as an abstract interlingua is Dedukti \cite{assaf-al-2023}.
Dedukti is meant to serve as a hub for translating between different proof systems and thereby enabling them to share code.
In addition, it provides an independent way of checking proofs in other system.
Dedukti itself is a maximally simple system, which minimizes the number of error sources.

Conversions to Dedukti are available from Agda, Lean, and Rocq, which are currently addressed by Informath, as well as to a growing number of other systems.
Conversions \textit{from} Dedukti are less common, because Dedukti aims to be strong enough to cover all the other systems and thereby stronger than any of them individually.
Translations from Dedukti are therefore deemed to be partial.
Translations from other systems \textit{to} Dedukti are, however, enough, if we are only interested in informalization.
And even if we want to perform formalization via Dedukti, we can often come a long way with partial translations.

What about using Dedukti as an interlingua for natural languages as well?
This is theoretically possible for any fragment of language that is produced by informalization from Dedukti.
However, doing this directly would destroy much of the simplicity and modularity we want to achieve.
Therefore, Informath uses \textit{two} interlinguas and mappings between them.
The other interlingua takes care of natural languages.
Informalization is defined as a mapping from Dedukti into this interlingua, and formalization is the inverse of this mapping.
The second interlingua and its natural language mappings are defined in Grammatical Framework (GF, \cite{ranta-2004,ranta-2011,ranta-al-2020}).

GF is a grammar formalism based on a logical framework similar to Dedukti \cite{ranta-2004}.
It was originally built as a translation system from constructive type theory \cite{martin-lof-1984} to natural language \cite{ranta-1994}.
Many of its applications have been on the language of mathematics, some of them multilingual \cite{hallgren-ranta-2000,ranta-2011c,caprotti-saludes-2012,kelber-al-2025}.

\subsection{Informath: a quick overview}

Informath has the following architecture:

\includegraphics[width=0.98\textwidth]{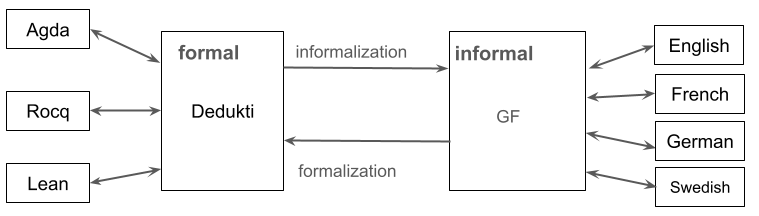}

\noindent
(Some of the arrows mark partial or one-to-many functions, as will be specified later.)

Here is an example statement involving the currently available languages.
The Dedukti statement has been used as the source of all the other formats. 
\begin{verbatim}
  Dedukti: prop110 : (a : Elem Int) -> (c : Elem Int) ->
    Proof (and (odd a) (odd c)) ->
    Proof (forall Int (b => even (plus (times a b) (times b c)))).
  
  Agda: postulate prop110 : (a : Int) -> (c : Int) ->
    and (odd a) (odd c) ->
    all Int (\ b -> even (plus (times a b) (times b c)))
  
  Rocq: Axiom prop110 : forall a : Int, forall c : Int,
    (odd a /\ odd c -> forall b : Int, even (a * b + b * c)) .
  
  Lean: axiom prop110 (a c : Int) (x : odd a /\ odd c) :
    forall b : Int, even (a * b + b * c)
\end{verbatim}

\begin{quote}
English: Prop110. Let $a$ and $c$ be integers. Assume that $a$ and $c$ are odd. Then $a b + b c$ is even for all integers $b$.

French: Prop110. Soient $a$ et $c$ des entiers. Supposons que $a$ et $c$ sont impairs. Alors $a b + b c$ est pair pour tous les entiers $b$.

German: Prop110. Seien $a$ und $c$ ganze Zahlen. Nimm an, dass $a$ und $c$ ungerade sind. Dann ist $a b + b c$ gerade für jede ganze Zahl $b$.

Swedish: Prop110. Låt $a$ och $c$ vara heltal. Anta att $a$ och $c$ är udda. Då är $a b + b c$ jämnt för alla heltal $b$.
\end{quote}

\noindent
The abstract syntax tree in GF notation looks as follows, in a visualization generated by GF software:

\includegraphics[width=0.98\textwidth]{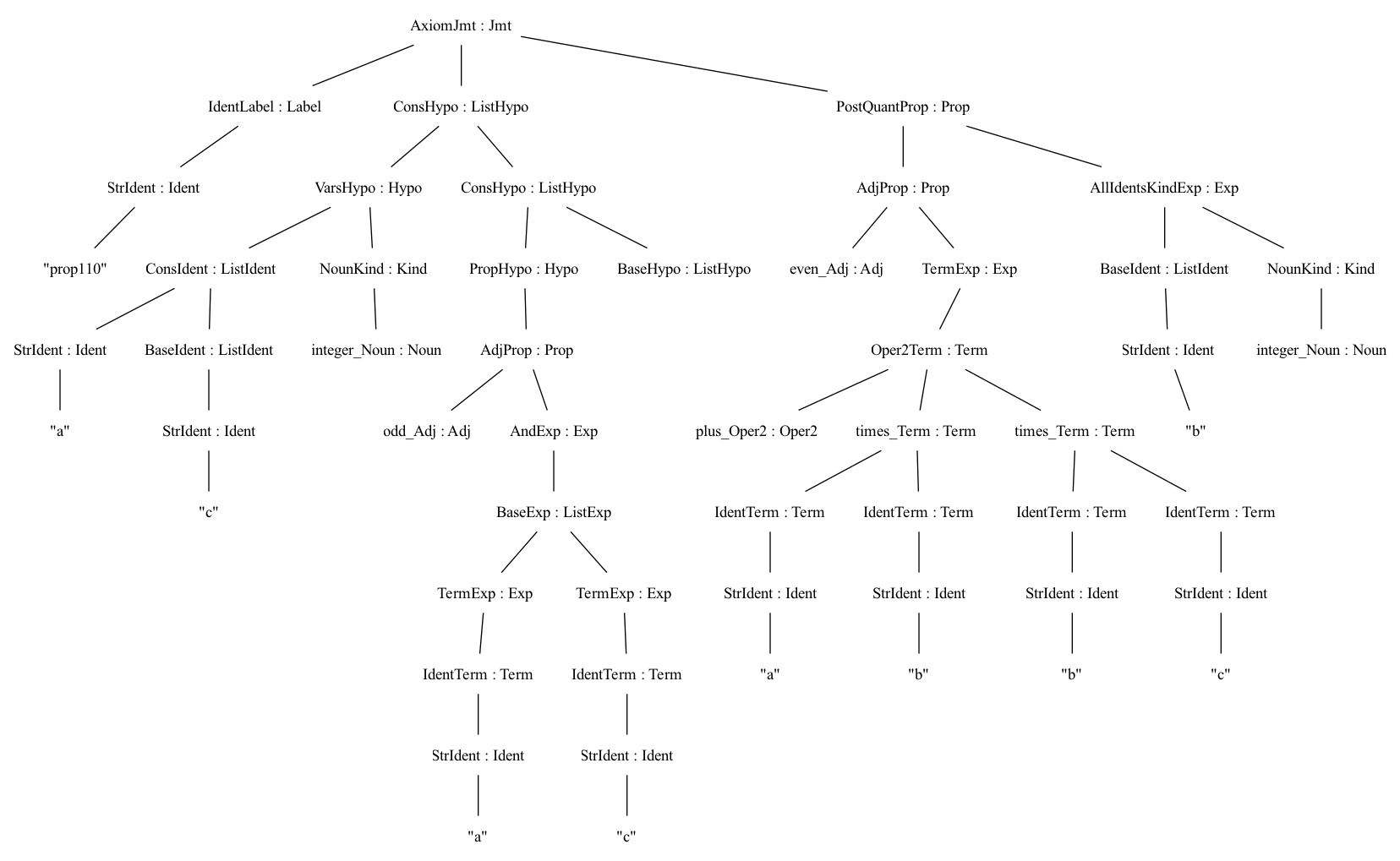}

\noindent
The GF tree is more complex than the Dedukti expression, because it adds grammatical information such as parts of speech (nouns, adjectives, etc.).
It is not aimed to be read by users, but it can be inspected at need, which is an essential part of the explainability of Informath.

\subsection{Dedukti}

\label{dedukti}

Dedukti is a minimalistic logical framework with dependent types and rewrite rules.
On the top level, a Dedukti module consists of \textbf{judgements} of three forms: \textbf{constant declarations} for either primitive or defined constants, and \textbf{rewrite rules}: 
\begin{verbatim}
  a : A.
  def a : A.
  [x, y, z ...] a --> b.
\end{verbatim}
In constant declarations, \texttt{A} is expected to be a Type. Type itself is technically not a Type, but a Kind, which is a meta-level notion not accessible from Dedukti code.

Rewrite rules can be used for definitions by cases, where $a$ is a \textbf{pattern} that matches certain expressions where the variables $x, y, z, \ldots$ may occur.
A special case is explicitly defined constants, with a single rewrite rule, 

\begin{verbatim}
  def a : A := b.
  \end{verbatim}
which is syntactic sugar for

\begin{verbatim}
  def a : A.
  [] a --> b.
  \end{verbatim}
(Dedukti also has judgement keywords \texttt{thm} and \texttt{inj}.
They differ from \texttt{def} only in how they are handled in computations and not in how they express mathematical content.)

Judgements are built from \textbf{expressions}, of some of the following forms:

\begin{compactitem}
\item identifier: \texttt{Ident}
\item application: \texttt{Exp Exp}
\item abstraction: \texttt{Ident =$>$ Exp}
\item dependent function type: \texttt{(Ident : Exp) -$>$ Exp}
\item non-dependent function type: \texttt{Exp -$>$ Exp}
\item local definition:  \texttt{(Ident : Exp := Exp) =$>$ Exp}
\end{compactitem}
Here is an example of how Dedukti has been used in Informath: the types \texttt{Set} and \texttt{Prop}, and the logical constants of a many-sorted predicate calculus: 

\begin{verbatim}
  Set : Type.
  Prop : Type.
  Elem : Set -> Type.
  Proof : Set -> Type.
  
  false : Prop.
  and : Prop -> Prop -> Prop.
  or  : Prop -> Prop -> Prop.
  if  : Prop -> Prop -> Prop.
  def not : Prop -> Prop := A => if A false.
  
  forall : (A : Set) -> (Elem A -> Prop) -> Prop.
  exists : (A : Set) -> (Elem A -> Prop) -> Prop.
  \end{verbatim}
As a framework based on dependent type theory, Dedukti also supports \textbf{proof objects}, which are typed by the propositions that they prove, in accordance with the propositions as types principle (the Curry-Howard isomorphism).
The following examples define natural deduction rules for implication in a way familiar from many logical frameworks:

\begin{verbatim}
  ifI : (A : Prop) -> (B : Prop) -> (Proof A -> Proof B) -> Proof (if A B).
  def ifE : (A : Prop) -> (B : Prop) -> Proof (if A B) -> Proof A -> Proof B.
  [c, a] ifE _ _ (ifI _ _ c) a --> c a.
  \end{verbatim}
If we want to analyse and generate standard informal mathematics, we also need to recognize \textbf{numeric literals}.
Since nothing about numbers is built in, there are many ways to define them.
This is one Dedukti definition we have used in Informath:

\begin{verbatim}
  Dig : Set.
  0 : Elem Dig.
  1 : Elem Dig.
  2 : Elem Dig.
  (; similarly for 3, 4, 5, 6, 7, 8, 9 ;) 
    
  Num : Set.
  nd : Elem Dig -> Elem Num.
  nn : Elem Dig -> Elem Num -> Elem Num.
  \end{verbatim}
In this encoding, $1987$ is expressed by 

\begin{verbatim}
  nn 1 (nn 9 (nn 8 (nd 7)))
  \end{verbatim}
The treatment of numbers (at least up to reals) as one type follows the analysis of numbers in informal mathematics in \cite{ganesalingam-2013}.
This simplifies the definitions of operators and predicates, where we can simply write

\begin{verbatim}
  plus : Num -> Num -> Num.
  less : Num -> Num -> Prop.
  \end{verbatim}
Finer type distinctions require coercions between types, usually in specific ways defined in translations between Dedukti and other formalisms.

\subsection{Type-correctness}

A fundamental notion in type theory is \textbf{type-correctness}: that an expression $a$ really has the type $A$.
By the propositions as types principle, type correctness also covers the correctness of proofs with respect to a given proposition.

Type correctness is defined by \textbf{typing rules}, which are implemented in a \textbf{type checker} program.
Typing rules are defined for judgements of the form
$$\Gamma \vdash a : A$$
where $\Gamma$ is a \textbf{context}, which assigns types to identifiers (i.e., variables and constants) and also contains the rewrite rules.
The type checker of Dedukti implements the following rules, which are standard for constructive type theory:
$$ \frac{\Gamma, x : A \vdash B : \text{Type}}{\Gamma \vdash (x : A) \rightarrow B : \text{Type}}$$

$$ \Gamma \vdash x : A, \text{if $x : A$ is in $\Gamma$}
\hspace{12mm}
\frac{\Gamma \vdash f : (x : A) \rightarrow B \hspace{4mm} \Gamma \vdash a : A}{\Gamma \vdash f a : B(x := a)} $$

$$ \frac{\Gamma, x : A \vdash b : B}{\Gamma \vdash x \Rightarrow b : (x : A) \rightarrow B}
\hspace{12mm}
\frac{\Gamma \vdash a : A \hspace{4mm} \Gamma \vdash A \equiv B}{\Gamma \vdash a : B}$$
The metanotation $B(x := a)$ means that $a$ is substituted for $x$ in $B$.
The metanotation $A \equiv B$ means that $A$ and $B$ are equal modulo rewrite rules.

The last rule makes type checking in Dedukti undecidable in general.
Since the rewrite rules form a Turing-complete language,  there is no general guarantee that their execution terminates.
Because of this, equality checking --- and hence type checking --- is undecidable.
In practice, this problem is avoided by careful design or by partial termination checkers.

\subsection{Grammatical Framework}

While Dedukti supports multilinguality on the formal side, GF does it on the informal side.
A \textbf{GF grammar} is a pair of an \textbf{abstract syntax} and a set of \textbf{concrete syntaxes}.
The abstract syntax has two forms of judgement: 

\begin{compactitem}
\item \textbf{category declaration}: \texttt{cat} $C$
\item \textbf{function declaration}: \texttt{fun} $f : C_1 \rightarrow \ldots \rightarrow C_n \rightarrow C$
\end{compactitem}
A concrete syntax is a mapping from the abstract syntax to linguistic objects such as strings and inflection tables.
It is defined by the following forms of judgement:

\begin{compactitem}
\item \textbf{linearization type}: \texttt{lincat} $C = T$, for each  \texttt{cat} $C$
\item \textbf{linearization function}: \texttt{lin} $f = t$, for each \texttt{fun} $f : C_1 \rightarrow \ldots \rightarrow C_n \rightarrow C$
\end{compactitem}
The simplest linearization type is \texttt{Str}, strings (more precisely, token lists).
But much of the power of GF comes from the use of more complex linearization types.
A linearization function must obey the linearization types of its argument and value types: if we denote the linearization type of $C$ by $C^*$ and the linearization function of $f$ by $f^*$, we must have
$$  f^* : C_1^* \rightarrow \ldots \rightarrow C_n^* \rightarrow C^* $$
An \textbf{abstract syntax tree} is an object built with applications of functions to arguments of appropriate types, their \textbf{subtrees}.
The \textbf{linearization} of an abstract syntax tree is an application of its main function to the linearizations of its subtrees:
$$  (f t_1 \ldots t_n)^* = f^* t_1^* \ldots t_n^* $$
Linearization is thus \textbf{compositional}, which is a limitation of what can be expressed in GF. 
It is an essential property of GF, used among other things in the conversion of linearization functions into parsing rules.

The multilinguality of GF comes from the possibility to define several concrete syntaxes for a single abstract syntax.
To give a very simple first example, consider the following grammar for propositions, expressions, and adjectives (actually a fragment of the full Informath grammar):

\begin{verbatim}
  cat Prop ; Exp ; Adj
  fun AdjProp : Adj -> Exp -> Prop
  fun even_Adj : Adj
  \end{verbatim}
A possible concrete syntax for English is

\begin{verbatim}
  lincat Prop, Exp, Adj = Str
  lin AdjProp adj exp = exp ++ "is" ++ adj
  lin even_Adj = "even"
  \end{verbatim}
This would not, however, take us very far: it can only cover predications with singular arguments, not ``$x$ and $y$ are even''.
The plural version could be achieved by using a richer linearization type, a \textbf{record} that indicates the number of the expression (singular or plural).
In this case, we can write

\begin{verbatim}
  lincat Exp = {s : Str ; n : Number}
  lin AdjProp adj exp = case exp.n of {
      Sg => exp.s ++ "is" ++ adj ;
      Pl => exp.s ++ "are" ++ adj
      }
  \end{verbatim}
\noindent
The type \texttt{Number} is defined as a \textbf{parameter type}, 

\begin{verbatim}
  param Number = Sg | Pl
  \end{verbatim}
In French, also the gender parameter needs to be added, because adjectives inflect for both number and gender; in German, also the case, and so on.
Since linearization types are defined in the concrete syntax of each language, the abstract syntax can be shared even if parameter systems differ.

\subsection{The Resource Grammar Library}

A crucial help for non-linguists using GF is the \textbf{Resource Grammar Library} (RGL).
The RGL implements the basic linguistic structures of around 40 languages (at the time of writing).
It has an API (application programming interface) that hides the actual linearization types and parameters.
With the RGL, the linearization rule of \texttt{AdjProp}\ can be simply written
\begin{verbatim}
  lin AdjProp adj exp = mkCl exp adj
\end{verbatim}
\vspace{-1mm}
which builds a clause (\texttt{Cl}) from a subject \texttt{exp} and a predicate \texttt{adj}.
The same RGL function \texttt{mkCl} is defined for all languages of the RGL, which have a wide range of different internal parameters and linearization types.

When it comes to lexical rules (rules for defining words), the RGL provides \textbf{smart paradigms}, which generate inflection tables from one or a few basic forms of words \cite{detrez-ranta-2012}.
For example, the French adjective paradigm \texttt{mkA} produces variations such as
\textit{pair, paire, pairs, paires} (just adding suffixes), 
\textit{commutatif, commutative, commutatifs, commutatives} (matching the final \textit{if}), and
\textit{réel, réelle, réels, réelles} (matching the final \textit{el}).

The RGL plays a major role in the productivity of Informath.
The abstractions usable in syntactic combination rules mean that the current set of 200 GF rules generate 87,000 context-free rules in the expansion of GF to BNF \cite{bringert-speechgram}.
Smart paradigms enable the extraction of a lexicon from a list of words with their parts of speech (noun, adjective, verb).

The RGL is also a major source of fluency: as it covers a wide variety of syntactic structures, it enables the extension of Informath to more and more constructs found in the ordinary language of mathematics.

\section{Productivity}

\label{productivity}

The productivity of symbolic methods in computational linguistics has been challenged on at least three grounds: labour intensity, knowledge requirements, and low coverage.
Their cost in money can hardly be seriously mentioned in this context any more: while trillions of dollars are spent on LLMs and they consume a substantial part of the Earth's resources, symbolic methods are being developed by a handful of people and usually run on average laptop computers.

If symbolic methods mean manually written context-free grammars and parsers, or the use of a sophisticated grammar formalism designed for linguists, they easily get either laborious or knowledge-intensive or both.
As claimed at the end of the previous section, much of this can be avoided by using GF, in particular the RGL.

But we suspect that most Informath users want more: they don't want to see GF at all, but just to use the system out of the box and at most give some hints about how their own concepts are to be verbalized.
Since the set of syntactic combinations already covers everything that is expressible in Dedukti, normal users should only need to tell how to deal with the concepts that are new in their projects.
This can be done in two ways:

\begin{compactitem}
\item by adding words to the GF grammar, which requires some knowledge of parts of speech and the target language;
\item by mapping formal concepts to the existing lexicon of Informath.
\end{compactitem}
\noindent
The ambition of Informath is that both of these tasks can be done by non-linguist proof system users \textit{without seeing any GF code}.

The implementation of Informath includes five kinds of code:

\begin{compactitem}
\item Dedukti (\texttt{.dk}), expressing the formal content,
\item GF (\texttt{.gf}), the grammar,
\item symbol tables (\texttt{.dkgf}), the mapping from Dedukti constants to GF,
\item Haskell (\texttt{.hs}), the overall system and conversions not definable in GF,
\item Python (\texttt{.py}), scripts for grammar extraction.
\end{compactitem}
Combining GF with Haskell is relatively smooth, via the technique of \textbf{embedded grammars}, which are Haskell datatypes generated from GF abstract syntax \cite{bringert-ranta-2008}.
But modifying the Haskell code is clearly a more involved task than just modifying the grammar, which in turn is more involved than writing symbol tables.
This hierarchy can be visualized as the \textbf{Informath deployment stack}:
\begin{center}
\includegraphics[width=0.64\textwidth]{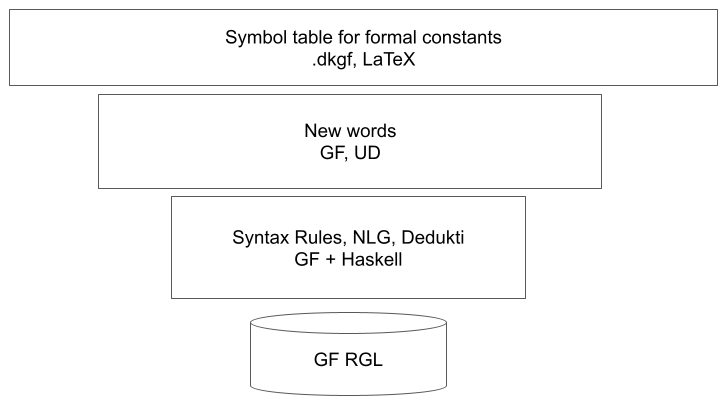}
\end{center}
\noindent
On the highest and the most commonly used level, a user who wants to adapt Informath to a new area of mathematics only needs to write a symbol table consisting of \textbf{lexical annotations} and \LaTeX\ macros, in a way to be explained in Section~\ref{symboltables}.
On the second-highest level, new words are added by GF smart paradigms, possibly with automated word collection based on techniques such as neural parsing with Universal Dependencies (UD \cite{demarneffe-al-2021}).
The second-lowest level needs GF code and usually some Haskell programming to link GF with Dedukti.
It is only needed when new syntactic forms and their semantic interpretations are added.
The lowest level is the GF RGL, which requires more serious GF programming and linguistic knowledge.
Informath, like most GF applications, can take this level for granted, except when ported to languages not yet supported by the RGL.

The following picture shows how the different files interact in the informalization of Dedukti code.
\begin{center}
\includegraphics[width=0.96\textwidth]{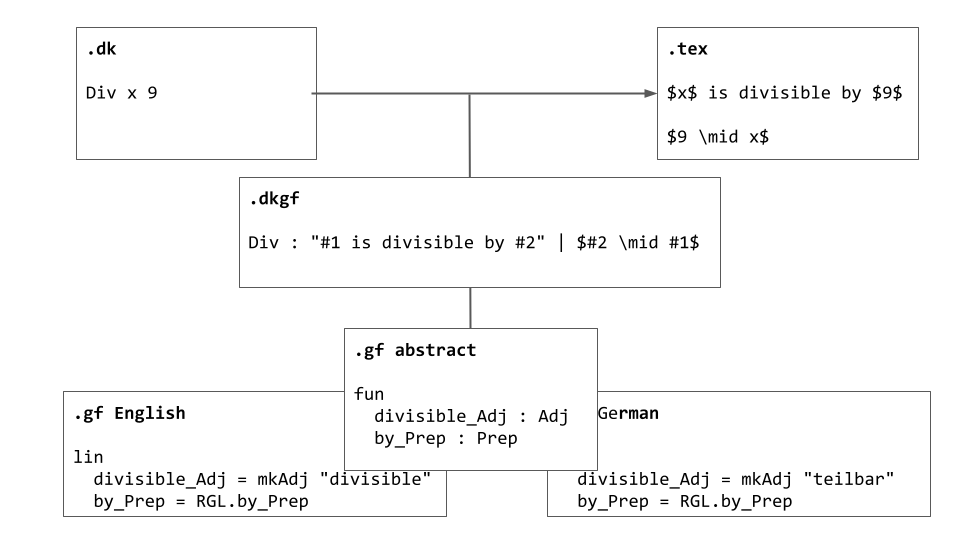}
\end{center}
\noindent

\section{Fluency}

\label{fluency}

Informath aims to support fluent and natural-sounding language similar to what is ordinarily used in mathematics.
This is more than usual CNLs, where it is enough to be semantically complete, unambiguous, and sufficiently close to natural language to be understandable without too much effort.

A thorough description of what ``natural-sounding'' means can be found in Ganesalingam's book \textit{The Language of Mathematics} \cite{ganesalingam-2013}.
Based on an analysis of several mathematical texts,  Ganesalingam concludes that we need

\begin{compactitem}
\item a mixture of words and symbols,
\item rich in grammatical structures,
\item enabling powerful abstractions,
\item syntactically ambiguous but always disambiguated by semantic clues.
\end{compactitem}
Since the language of mathematics is open-ended and therefore not precisely definable with a formal grammar, the ambition of Informath is limited to \textit{approximate} it.
A more accurate specification of the task is semantic and linguistic completeness and soundness:

\begin{compactitem}
\item \textbf{Semantic completeness}: everything expressible in type theory (Dedukti) can be translated with Informath without loss of meaning. This is achieved by a total function from Dedukti to Informath abstract syntax. 
\item \textbf{Semantic soundness}: everything expressible in Informath can be translated to Dedukti without loss of meaning. This is achieved by a total function from Informath abstract syntax to Dedukti.
\item \textbf{Linguistic completeness}: whenever a form of expression is attested in mathematical texts, it should be eventually added to Informath. This is approached by gradually adding new grammar rules when needed to parse a text.
\item \textbf{Linguistic soundness}: nothing should be added to Informath unless it is attested in mathematical texts. This is approached by generating new texts and assessing them by human evaluation (slow but precise) or by language models (fast but less precise).
\end{compactitem}
\subsection{The structure of the grammar}

The grammar of Informath is divided into two main parts:

\begin{compactitem}
\item MathCore, a minimalistic CNL, designed to cover all of Dedukti in an almost one-to-one fashion,
\item MathExtensions, an extension of MathCore aiming to approximate the full language of mathematics.
\end{compactitem}
The relations between MathCore and MathExtensions are shown in the following picture:

\begin{center}
\includegraphics[width=0.72\textwidth]{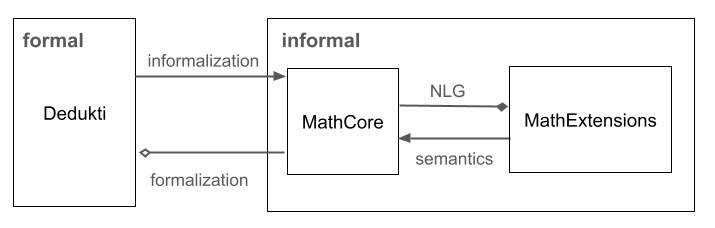}
\end{center}

\noindent
This picture uses solid arrowheads to say that the operation is a total function, giving either exactly one result for every input (triangular arrowheads) or possibly many (diamond).
Hollow arrowheads mean partial functions, which can likewise give at most one result (triangular) or many results (diamond).
Thus the conversion from MathCore to Dedukti may fail because MathCore is more permissive than Dedukti; this is because Informath delegates dependent type checking to Dedukti.
This conversion can, moreover, give multiple results, because of overloaded words and symbols.

The division of the grammar to core and extensions aims to separate concerns.
All conversions from Dedukti and back take place in MathCore, which takes care of semantic soundness and completeness.
These conversions are defined in Haskell, by using the technique of embedded GF grammars \cite{bringert-ranta-2008}.
They are relatively stable, because MathCore is already complete and not expected to change.
Its changing part is the lexicon, but it is related to Dedukti by symbol tables that are loaded at runtime; see Section~\ref{symboltables}.

The purpose of MathExtensions is to approximate linguistic completeness.
Linguistic soundness is delegated to the RGL, for the aspects that concern grammatical correctness.
Soundness in the stricter sense of attest in mathematical texts is a responsibility of Informath developers.

The semantic soundness of MathExtensions is established by a conversion to MathCore.
This conversion is similar to logical semantics of Montague style \cite{montague-1973}.
It is a recursive function on abstract syntax trees defined in Haskell.
When new functions are added to the grammar, corresponding semantic rules must be added to the Haskell code; in certain simple cases they can be read from symbol tables (Section~\ref{symboltables}).

The conversion from MathCore to MathExtensions, named NLG (\textbf{Natural Language Generation}) in the picture, is one-to-many.
It always results in at least one value, the MathCore expression itself.
To guarantee semantic soundness, it should be an inverse of the semantics: for all trees $t$ in MathCore,
$$
\text{NLG}(t) \subseteq \{t'  \,\mid\, \text{semantics}(t') = t\}
$$
where ``NLG'' and ``semantics'' refer to the above diagram.
One might expect an equality of these sets --- that the NLG produces exactly those trees that semantically valid variants of the MathCore tree.
But we have found this to be too much: to guarantee fluency and avoid misleading verbalizations, the result of NLG should be just a subset of what the semantics would allow.
Even with such contraints, the NLG component of Informath often produces thousands of variants for every Dedukti judgement.

\subsection{Natural language generation}

\label{nlg}

The fluency of Informath is based on the classical theory of NLG \cite{reiter-dale-2000}.
It this theory, NLG consists of the following steps:

\begin{compactitem}
\item \textbf{Content determination}: what to say.
\item \textbf{Text planning}: in what order to say it.
\item \textbf{Aggregation}: merging atomic facts into compound sentences.
\item \textbf{Lexical choice}: selecting words for concepts.
\item \textbf{Referring expression generation}: using pronouns and other alternative expressions for entities.
\item \textbf{Surface realization}: selecting the final word order and morphological forms.
\end{compactitem}
Linearization of abstract syntax trees in GF corresponds to surface realization, whereas lexical choices are defined by symbol tables.
Content determination is ultimately done by the Dedukti formula, but it can also involve the omission of details, in particular in the informalization of proofs, which also require text planning. 

The goal of NLG is to made verbalizations compact and fluent, while avoiding ambiguity. 
The way this is done in Informath is to generate variant expressions and rank them with several criteria taken into account.
The current ranking is based on a penalty, which is a linear combination of the following counts, which take into account both the tree and its linearization:

\begin{compactitem}
\item tree size,
\item tree depth (favouring shallow structures),
\item number of characters,
\item number of tokens,
\item subsequent dollars (the style rule that symbolic parts, marked by dollar characters in \LaTeX, should be separated by words \cite{knuth-al-1987}),
\item initial dollars (the style rule that a sentence should start with a word and not symbols),
\item alternative parses (favouring unambiguous expressions).
\end{compactitem}
The weights of each count can be set by the user, but they are also a candidate for machine learning.
An alternative way of ranking could be to use language models, as in \cite{le-zhou-2022}.
Since all variants generated by NLG rules are known to be semantically correct, the LLM cannot introduce ``hallucinations''.

The implementation of NLG is a pipeline similar to compiler optimizations, as observed in \cite{coscoy-al-1995} (without using the term NLG).
It can be structured to different passes, such as aggregation, but it can be even more fine-grained.
The actual NLG code in Informath currently consists of ten passes, each of which uses the tecnnique of \textbf{almost compositional function}, which enable the threading of recursive functions over large datatypes (such as the abstract syntax of Informath) by treating explicitly only the cases that are not compositional \cite{bringert-ranta-2008}.
This has proven to be a productive technique to improve the fluency of NLG with GF.
It also allows the user to select and rank subsets of NLG steps, over and above the general ranking shown above.

\section{A tour of the Informath grammar}

\label{atour}

In this long section, we will go through the main linguistic structures covered by the current grammar of Informath.
We will often start with MathCore, because it is the basis of the informalization of Dedukti code.
But we will not always make a clear distinction between MathCore and MathExtensions, because it is a somewhat arbitrary technical detail that does not say much about the ordinary language of mathematics.

\subsection{Syntactic categories}

\label{SyntacticCategories}

The syntactic category system of Dedukti is extremely simple: the topmost level is modules, which are sequences of judgements, which are built from expressions, whose ultimate parts are identifiers.
The natural language grammar of Informath is considerably richer, as it tries to cover all of the ways in which mathematics is ordinarily expressed.

Most of the variation concerns Dedukti's category of expressions (\texttt{Exp}), which can be informalized as noun phrases, sentences, or even as entire texts.
Expressions consisting of natural language words will be called \textbf{verbal}.
In addition to them, the language of mathematics has \textbf{symbolic} expressions, which consist of mathematical symbols\footnote{This sense of "symbolic" is different from "symbolic" as opposed to "statistical" or "neural" used in the title of this paper.}. 
The following table lists the current Informath categories corresponding to \texttt{Exp}, together with linguistic categories (similar to linearization types implemented in the RGL) and examples of each.
The first four categories are verbal, the last two symbolic.

\begin{tabular}{@{}llll@{}}
category & name & linguistic type & example \\
\midrule
\texttt{Exp} & expression & noun phrase & \textit{the empty set} \\
\texttt{Kind} & kind & common noun & \textit{rational number} \\
\texttt{Proof} & proof & proof text & \textit{By Theorem 3, p is prime.} \\
\texttt{Prop} & proposition & sentence & \textit{p is prime} \\
\texttt{Formula} & formula & formula & $x + y = z$ \\
\texttt{Term} & term & term & $x + y$ \\
\end{tabular}

\noindent
Each of these categories has several abstract syntax functions to build phrases.
Thus noun phrases can be build from common nouns with determiners and possible arguments (\textit{the successor of $n$}), common nouns can be modified with adjectives (\textit{even number}), and sentences can be built with adjectives (\textit{$n$ is even}) and verbs (\textit{$l$ intersects $m$}), just to give some examples.
Verbal expressions can contain parts that are symbolic, by conversion functions of the GF types

\begin{compactitem}
\item \texttt{Formula -$>$ Prop}
\item \texttt{Term -$>$ Exp}
\end{compactitem}
whereas symbolic expressions are usually built only from symbolic parts; the most common exception is set comprehensions.

Natural language also has the versatility to convert between verbal categories.
The following conversions are included in the Informath grammar and can be combined; the argument of each conversion is shown in boldface:

\begin{compactitem}
\item \texttt{Exp -$>$ Kind}, e.g. \textit{element of \textbf{the empty set}}
\item \texttt{Kind -$>$ Exp}, e.g. \textit{the set of \textbf{rational numbers}}
\item \texttt{Prop -$>$ Kind}, e.g. \textit{proof that \textbf{p is prime}}
\item \texttt{Kind -$>$ Prop}, e.g. \textit{there is a \textbf{rational number}}
\item \texttt{Exp -$>$ Proof}, e.g. \textit{by \textbf{the axiom of choice}}
\item \texttt{Proof -$>$ Exp}, e.g. \textit{the following proof: \textbf{assume} \ldots}
\end{compactitem}
These conversions are sometimes needed in the informalization of well-typed Dedukti statements, especially ones relying on the propositions as types principle.
They must therefore be complete, even though they can produce clumsy language.
For instance, the chain leading from a \texttt{Prop} $P$ to \texttt{Exp} is \textit{the set of proofs that} $P$.
The conversions can often be avoided by giving alternatives of different linguistic types in symbol tables.

\subsection{Constant applications and lexical categories}

The basic and most open-ended level of the Informath grammar are expressions that correspond to atomic formulas and function applications.
In Dedukti syntax, they are \textbf{constant applications}, expressions of the form
$$ c \, a_1 \ldots a_n $$
where
$$ c : (x_1 : C_1) \rightarrow \ldots \rightarrow (x_n : C_n) \rightarrow C $$
is a constant declared in a judgement and $a_1, \ldots, a_n $ are expressions of the corresponding argument types of $c$.
Any of the bound variables $x_i$ may be omitted, if the rest does not depend on it.
We assume $C$ to be a ground type (not a function type), thus considering \textbf{total applications}, where the constant $c$ takes the maximum number of arguments that its type permits.

Choosing natural, widely used verbalizations for constant applications is a crucial component of fluent language.
To support the different forms of expression that occur in mathematical language, Informath provides several \textbf{lexical categories}.
Table~\ref{lexcat} displays a set of categories with their ``semantic types'', which indicate the linguistic types of arguments and value.
It also shows examples of linearizations in English.
They are shown in a format that can be used in the lexical annotations of Dedukti constants.

\begin{table}

\begin{tabular}{@{}lll@{}}
category & semantic type & example \\
\midrule
\texttt{Adj} & \texttt{Exp -$>$ Prop} & \textit{X is even} \\
\texttt{Adj2} & \texttt{Exp -$>$ Exp -$>$ Prop} & \textit{X is divisible by Y} \\
\texttt{Adj3} & \texttt{Exp -$>$ Exp -$>$ Exp -$>$ Prop} & \textit{X is congruent to Y modulo Z} \\
\texttt{AdjC} & \texttt{Exps -$>$ Prop} & \textit{X and Y are distinct} \\
\texttt{Dep} & \texttt{Exp -$>$ Kind} & \textit{root of X} \\
\texttt{Dep2} & \texttt{Exp -$>$ Exp -$>$ Kind} & \textit{interval from X to Y} \\
\texttt{DepC} & \texttt{Exps -$>$ Kind} & \textit{path between X and Y} \\
\texttt{Fam} & \texttt{Kind -$>$ Kind} & \textit{list of As} \\
\texttt{Fam2} & \texttt{Kind -$>$ Kind -$>$ Kind} & \textit{function from As to Bs} \\
\texttt{Fun} & \texttt{Exp -$>$ Exp} & \textit{the square of X} \\
\texttt{Fun2} & \texttt{Exp -$>$ Exp -$>$ Exp} & \textit{the image of X in Y} \\
\texttt{FunC} & \texttt{Exps -$>$ Exp} & \textit{the sum of X and Y} \\
\texttt{Label} & \texttt{Proof} & \textit{theorem 1 .} \\
\texttt{Name} & \texttt{Exp} & \textit{the empty set} \\
\texttt{Noun} & \texttt{Kind} & \textit{integer} \\
\texttt{Noun1} & \texttt{Exp -$>$ Prop} & \textit{X is a prime} \\
\texttt{Noun2} & \texttt{Exp -$>$ Exp -$>$ Prop} & \textit{X is a divisor of Y} \\
\texttt{NounC} & \texttt{Exps -$>$ Prop} & \textit{X and Y are relative primes} \\
\texttt{Verb} & \texttt{Exp -$>$ Prop} & \textit{X converges} \\
\texttt{Verb2} & \texttt{Exp -$>$ Exp -$>$ Prop} & \textit{X divides Y} \\
\texttt{VerbC} & \texttt{Exps -$>$ Prop} & \textit{X and Y coincide} \\
\end{tabular}

\caption{
Lexical categories of verbal expressions with semantic types and examples.
The category \texttt{Exps} contains non-empty lists of expressions, where the last two expressions are combined with the conjunction ``and'' and its equivalent in different languages.
The letter \texttt{C} marks collective relations and functions.
}
\label{lexcat}

\end{table}

\noindent

As the examples in table~\ref{lexcat} shows, functions of \texttt{C}-marked categories can take their arguments as \textit{and}-separated lists.
These lists have typically length 2, but they can also be longer.
They can sometimes be reduced to conjunctions of 2-place predications.
For instance, \textit{$a, b$ and $c$ are equal} means the same as \textit{$a$ and $b$ are equal and $b$ and $c$ are equal}.
In contrast, ``linearly independent'' is not analysable in this way: the sentence \textit{all vectors in $X$ are linearly independent} is an irreducible example of \texttt{AdjC} \cite{arambillete-degroote-2025}.
Similar considerations apply to \texttt{FunC}: it is sometimes reducible to ``folding'' operations of two-place functions, as in \textit{the sum of $a, b$ and $c$}.
Sometimes it is not, as in \textit{the plane determined by $a, b$ and $c$}.
Such distinctions can be expressed in symbol tables, and their correctness should be checked with respect to the formal code. 

\subsection{Complex lexical items}

Lexical items are verbalizations of atomic Dedukti constants.
But they need not be single words themselves: on the contrary, mathematical terminology is full of \textbf{multiword expressions}.
The \texttt{Adj} category of Informath thus corresponds to \textbf{adjectival phrases} rather than just single-word adjectives, and the same applies to \texttt{Noun} and \texttt{Verb}.
Here are some productive ways to form adjectives, nouns, and verbs, shown as GF abstract syntax functions with an example of each function:

\begin{verbatim}
  AdverbAdjAdj : Adverb -> Adj -> Adj             -- uniformly continuous
  AdjNounNoun : Adj -> Noun -> Noun               -- real number
  NounNounNoun : Noun -> Noun -> Noun             -- vector space
  ProperNameNounNoun : ProperName -> Noun -> Noun -- Hilbert space      
  \end{verbatim}
These rules permit the formation of arbitrarily long lexical items, such as \textit{metric Banach manifold}, whose abstract syntax is

\begin{verbatim}
  AdjNounNoun metric_Adj (ProperNameNounNoun Banach_ProperName manifold_Noun)
\end{verbatim}
As shown in these examples, the abstract syntax functions have systematic names, trying to make them easy for users to remember and even guess.

Similar rules permit the formation of functions of complex categories from simpler ones, for example,

\begin{verbatim}
  AdjPrepAdj2 : Adj -> Prep -> Adj2  -- distinct from
  AdjAdjC : Adj -> AdjC              -- distinct (collective)
  \end{verbatim}
Thanks to structured lexical items, multiword items do not need to be included in the grammar, but they can be built from simple words in symbol tables, as described in the next section.
Notice, however, that multiword expressions are often not translated compositionally in different languages.
For instance, English may use compounds such as \textit{vector space} where French uses adjectival modifiers as in \textit{espace vectoriel}.
This is not a crucial problem, because the verbalization of a Dedukti constant can be defined in separate symbol tables for different languages.

\subsection{Lexical annotations and symbol tables}

\label{symboltables}

Lexical annotations are entries in \textbf{symbol tables}, which are read at runtime from files with suffix \texttt{.dkgf}.
They can refer to abstract syntax functions in the existing GF lexicon, such as

\begin{verbatim}
  NEq : distinct_Adj2 | distinct_AdjC
  \end{verbatim}
which generates the variants \textit{X is distinct from Y} and \textit{X and Y are distinct}.

Lexical annotations in symbol tables can also use quoted expressions that are parsed by using the lexicon.
The above entry could equally be written as

\begin{verbatim}
  NEq : "X is distinct from Y" | "X and Y are distinct"
  \end{verbatim}
These expressions can be parsed in the categories \texttt{Adj2} and \texttt{AdjC}, respectively.
They do not require there to be separate entries for \texttt{distinct\_Adj2} and \texttt{distinct\_AdjC} in the lexicon: it is enough that there is \texttt{distinct\_Adj}.
The parse result is in that case a complex lexical item.

Symbol table entries for categories whose expressions take arguments use a fixed set of variables, marked with \texttt{X}, \texttt{Y} or \texttt{Z} for \texttt{Exp} arguments and with \texttt{A} or \texttt{B} for \texttt{Kind} arguments.
The symbol table also specifies prepositions for each argument place, including no preposition, as for instance with transitive \texttt{Verb2} verbs.

An alternative format for variables in symbol table entries is similar to what is used in \LaTeX\ macro definitions: \verb:#1:, \verb:#2:, etc.
The numbers refer to argument places, which makes it possible to express permutations:
\begin{verbatim}
  Div : "#1 divides #2" | "#2 is divisible by #1"
\end{verbatim}
This notation also enables dropping argumens, which is often needed in informalization.
The type-dependent equality predicate is a prime example: its Dedukti typing is
\begin{verbatim}
  Eq : (A : Set) -> Elem A -> Elem A -> Prop.
\end{verbatim}
The symbol table entry omits the first argument:
\begin{verbatim}
  Eq : "#2 is equal to #3" | "#2 and #3 are equal"
\end{verbatim}
An equivalent way to omit a specific number of arguments from the beginning of the list is \texttt{\#DROP} directives:
\begin{verbatim}
  #DROP Eq 1
  Eq : "X is equal to Y" | "X and Y are equal"
\end{verbatim}
However, this method is less general than numbered arguments, because it only applies to initial arguments and because it does not permit dropping different arguments in different alternatives for a given Dedukti constant.

\subsection{Symbol table extensions}

Symbol tables were originally designed to verbalize atomic expressions with lexical items, which are the most important way in which users need to extend Informath.
This was later generalized to arbitrary GF functions that do not belong to lexical categories, such as ones for logical constants:

\begin{verbatim}
  and : CoreAndProp  # GF type Prop -> Prop -> Prop
  \end{verbatim}
From this perspective, the purely lexical annotations are a special case, in fact shorthand for \textbf{lexical application functions} applied to lexical functions but not yet to arguments.
For example, when we write

\begin{verbatim}
  NEq : AdjPrepAdj2 from_Prep distinct_A
  \end{verbatim}
or its equivalent \texttt{"X and Y are distinct"}, the GF function assigned to \texttt{NEq} is the partial application

\begin{verbatim}
  Adj2Prop (AdjPrepAdj2 from_Prep distinct_A)
  \end{verbatim}
of the lexical application function of 2-place adjectives,

\begin{verbatim}
  Adj2Prop : Adj2 -> Exp -> Exp -> Prop
  \end{verbatim}
The effect is the following translation rule for Dedukti expressions, where $t^*$ denotes the translation of $t$:
$$
(\text{NEq} \, x \, y)^* = \text{Adj2Prop (AdjPrepAdj2 from$\_$Prep distinct$\_A$)} \, x^* y^*
$$
To summarize: if the type of a symbol table entry is a lexical category, the lexical application function is added to it.
Otherwise, the assigned GF expression is used as is.

Symbol tables are type-checked by comparing the type of the Dedukti function with the type of the GF function: they must have the same arities and the syntactic types of their arguments must match.
There is, moreover, a restriction on the kinds of GF functions that can be used: they must be either constants or applications of constants.
Lambda abstractions are not allowed, because they can be impossible to apply backwards when converting Informath to Dedukti.

The use of example strings in symbol tables needs a word of explanation.
These strings are parsed in a special category \texttt{Example}, with functions such as

\begin{verbatim}
  Adj2Example : Adj2 -> Argument -> Argument -> Example
  \end{verbatim}
For instance, \texttt{"X is distinct from Y"} is parsed as

\begin{verbatim}
  Adj2Example (AdjPrepAdj2 from_Prep distinct_A) X_Argument Y_Argument
\end{verbatim}
From such a tree, it is easy to extract the GF function to be used: it is the first subtree.
The parser is able to identify the intended lexical category as \texttt{Adj2}, in contrast to \texttt{Adj}\ due to the presence of the two arguments, and to \texttt{AdjC}\ due to the preposition ``from''.

Yet another extension of symbol tables allows compositional semantic rules to convert new syntactic functions to MathCore, such as
\begin{verbatim}
  #SEMANTICS InverseIfProp b a = IfProp a b
  \end{verbatim}
where \texttt{InverseIfProp} enables the expression of \textit{if A then B} as \textit{B if A}.
In this way, a user can add not only lexical items but also syntactic combinations to the grammar without modifying and recompiling the Haskell code of Informath.
Similar directives can be defined for NLG to generate variant expressions in a user-controllable way:
\begin{verbatim}
  #NLG IfProp a b = InverseIfProp b a
\end{verbatim}
Notice that \verb6#NLG6 directives are expected to obey the semantics, but they are not automatically derived from semantics, because we want the user to be able to control what semantic rules are applied in the NLG direction.

In addition to GF functions, symbol tables enable definitions of \LaTeX\ macros for symbolic expressions, to be described in Section~\ref{symbolic}.
The general purpose of symbol table extensions is to move functionalities upwards in the Informath deployment stack; macros, in particular, make it possible to define new symbolic notations without extending the grammar.

\subsection{Using the lexicon in translation}

As shown above, the symbol table used in converting between Dedukti and GF assigns to each Dedukti constant $c$ a set of GF functions.
Let us denote this set with $c^{G}$ and the translation set of an expression $e$ with $e^{*}$.
We can then define the set of translations of the application of a Dedukti constant $c$ to GF as follows:
$$ (c \, a_1 \ldots a_n)^{*} = \{ f \, b_1 \ldots b_n \mid f \in c^{G}, b_1 \in a_1^{*}, \ldots , b_n \in a_n^{*} \} $$

\noindent
The opposite direction of translating Informath to Dedukti uses the same symbol table but in reverse.
It can also give several results, but now with a different reason: natural language lexical functions may be \textbf{overloaded}, as one and the same lexical function can verbalize different Dedukti constants.
Denoting the inverse translation with $e^{-1}$ and the set of all Dedukti constants with $D$, we can define the translation of the application of a GF constant $f$ to Dedukti as follows: 
$$ (f \, b_1 \ldots b_n)^{-1} = \{ c \, a_1 \ldots a_n \mid c \in D, f \in c^{G}, a_1 \in b_1^{-1}, \ldots , a_n \in b_n^{-1} \} $$
In such cases, one can normally expect all of the resulting constants to have different Dedukti types, so that the type checker of Dedukti can resolve the overloading.
Such overloading is quite common in ordinary mathematical language, as shown in \cite{ganesalingam-2013}.
In addition, dropped arguments must be restored.
This can be initialized by inserting metavariables and completed by resolving them as a part of type checking.

\subsection{Symbolic notations}

\label{symbolic}

The generation of symbolic notation in Informath operates on two different categories: formulas (\texttt{Formula}) corresponding to propositions (\texttt{Prop}), and terms (\texttt{Term}) corresponding to expressions (\texttt{Exp}).
In the traditional style of informal mathematics, only atomic propositions have symbolic notations --- not logically complex ones. 
But if the object language of informalization is formal logic, it is of course possible to define complex formulas as expressions of the category \texttt{Term}.

Symbolic notations can be defined in a fine-grained way by the following syntactic categories:

\vspace{2mm}
\begin{tabular}{@{}lll@{}}
category & semantic type & example \\
\midrule
\texttt{Compar} & \texttt{Term -$>$ Term -$>$ Formula} & \texttt{$<$} \\
\texttt{Const} & \texttt{Term} & \texttt{$\backslash$pi} \\
\texttt{Oper} & \texttt{Term -$>$ Term} & \texttt{$\backslash$sqrt} \\
\texttt{Oper2} & \texttt{Term -$>$ Term -$>$ Term} & \texttt{+} \\
\end{tabular}
\vspace{2mm}

\noindent
For example, the Dedukti functions

\begin{verbatim}
  Neq : Elem Num -> Elem Num -> Prop
  plus : Elem Num -> Elem Num -> Elem Num  
  pow : Elem Num -> Elem Num -> Elem Num
  \end{verbatim}
with the symbol table

\begin{verbatim}
  Neq : Neq_Compar
  plus : plus_Oper2
  pow : pow_Oper2
  \end{verbatim}
permits the informalization of the Dedukti expression

\begin{verbatim}
  Neq (plus (pow a n) (pow b n)) (pow c n)
  \end{verbatim}
as the symbolic formula

\begin{verbatim}
  $a^{n} + b^{n} \neq c^{n}$
  \end{verbatim}
which \LaTeX\ renders as
$$ a ^ {n} + b ^ {n} \neq c ^ {n} $$
provided that the linearization of \verb:Neq_Compar: uses the infix operator $\neq$, and so on.

In addition to notations defined in the grammar, Informath supports the definition  of arbitrary \textbf{macros} in symbol tables.
A macro definition can be included in a symbol table entry as follows:
\begin{verbatim}
  congruent : "X is congruent to Y modulo X" | $#1\equiv#2\,\text{mod}\,#3$
\end{verbatim}
Thus, symbolic renderings are surrounded by \verb6$6 characters to distinguish them from verbal renderings surrounded by quotes.
In the translation to text, this piece of \LaTeX\ code is translated to a macro definition,
\begin{verbatim}
  \newcommand{\congruent}[3]{#1\equiv#2\,\text{mod}\,#3}
\end{verbatim}
which is copied to generated \LaTeX\ documents and thereby made usable in informalization.
A drawback is that the direction of formalization then only works for \LaTeX\ code that is written by using the declared macros.
The informalization direction has an issue as well: precedence.
Macros give no way to control the use of parentheses in the expressions that are generated.
The cautious solution used in the Informath grammar is to define the linearization of macro expressions so that enough parentheses are always added.
This may lead to expressions that have more parentheses than in natural mathematical text.
If this is not tolerated, the user can move one step down the Informath deployment stack and define GF functions with desired precedences.

Just like verbal expressions, the generation of symbolic expressions is driven by symbol tables. 
If we denote the set of symbolic functions (complete with lexical application functions) for a constant $c$ by $c^{F}$ and the set of symbolic variants of an expression $e$ by $e^{S}$, we can define the set of symbolic formulas and terms concisely:

$$ (c \, a_1 \ldots a_n)^{S} = \{  f \, s_1 ... s_n  \mid f \in c^{F}, s_1 \in a_1^{S}, \ldots, s_n \in a_n^{S}  \}$$
\noindent
A built-in restriction of this definition is that symbolic terms (and formulas) can only have symbolic parts. 
It does not cover symbolic expressions with verbal parts, in particular comprehensions:

$$ \{x \in \mathbb{N} \mid \text{$x$ is even and prime} \} $$
\noindent
which must be treated separately.
The reason why this form is needed is that there are propositions that can only be expressed by verbal means --- the predicates ``even'' and ``prime'' in usual textbook mathematics are examples of this, as well as the conjunction ``and''.

Even symbolic notations can be overloaded, which creates ambiguities in the inverse translation.

\subsection{Variable-binding constants}

The standard way of dealing with variable-binding expressions such as quantifiers, sums, and integrals, in type theory is with functions that take functions as their arguments.
For example, the integral expression
$$
\int_a^b f(x) dx
$$
can be formalized with the constant

\begin{verbatim}
  integral : (a : Elem Real) -> (b : Elem Real) ->
     (f : ((x : Elem Real) -> Elem Real)) -> Elem Real.
\end{verbatim}
Thanks to this encoding, the type theory only needs to deal with one variable-binding construct, the function abstract.
The grammatical term for this method is \textbf{higher-order abstract syntax} (HOAS), because it uses higher-order functions to represent syntactic combinations.

Informath has some support for user-defined lexical annotations corresponding to HOAS.
Thus the above Dedukti function can be given the symbol table entry
\begin{verbatim}
  integral : "the definite integral of Z where $x$ ranges from X to Y"
      | $\int_{#1}^{#2} #4 d#3$
\end{verbatim}
The arguments \texttt{X} and \texttt{Y} correspond to the limits \texttt{a} and \texttt{b}, respectively, whereas the argument \texttt{Z} corresponds to the body of the function \texttt{f}, and the argument \texttt{\$x\$} (given in \LaTeX\ dollar signs) corresponds to the bound variable \texttt{x} in the Dedukti type.

In the symbolic expression for integrals, the numbering of the arguments corresponds to flattening the Dedukti function into first order, where the variable occurs before the body of the function argument; this is performed as a preprocessing step in informalization:
\begin{verbatim}
  integral : (a : Elem Real) -> (b : Elem Real) ->
      (x : Ident) -> (f : Elem Real) -> Elem Real.
  \end{verbatim}
The treatment of constants of higher-order types as variable-binding operators works well with much of traditional undergraduate mathematics.
However, it gives unwanted informalizations in more advanced mathematics where function arguments are meant to be function objects.
An experiment with homotopy type theory \cite{rijke-2025} produced examples of this \cite{ohlsson-ranta-2026}.
The adapted solution was to define a new type constructor, \texttt{Map}, for functions used as objects:

\begin{verbatim}
  def Map : (A : U) -> (B : (El A -> U)) -> Type := A => B => (x : A) -> B x.
\end{verbatim}
With this, the type of homotopies can be defined without explicit higher-order functions

\begin{verbatim}
  htpy : (A : U) -> (B : (El A -> U)) -> Map A B -> Map A B -> Type.
\end{verbatim}
and verbalized with the symbol table entries

\begin{verbatim}
  #DROP htpy 2
  htpy : "homotopy between X and Y" | $#1 \sim #2$
\end{verbatim}
which produces symbolic expressions of forms $f \sim g$.

\subsection{Logical constants}

Lexical categories and their application functions cover atomic propositions, singular terms, and basic types.
Moving upwards from that level, we have to deal with logically complex propositions and with judgements.
These structures are lower down the Informath deployment stack (Section~\ref{productivity}), which means that it is harder to change them or add new ones.
On the other hand, the already existing combinations are sufficient to express everything in Dedukti, usually in numerous different ways.

We can assume that logical operations are expressed in Dedukti with the constants listed in Section~\ref{dedukti}; this behaviour can be changed by modifying the symbol table.
The baseline translation of logical formulas is familiar form many CNLs:

\begin{verbatim}
  and A B           ==> A and B
  or A B            ==> A or B
  if A B            ==> if A then B
  not A             ==> it is not the case that A
  forall A (x => B) ==> for all x in A, B
  exists A (x => B) ==> there exists an A x such that B
  \end{verbatim}
These schematic rules encode several steps.
By using the first rule as example, they work as follows:
To translate a Dedukti expression of form \texttt{and A B} into a tree of type \texttt{Prop}

  \begin{enumerate}
  \item Translate \texttt{A} and \texttt{B} into trees of type \texttt{Prop}.
  \item Combine these trees with an abstract syntax function \texttt{CoreAndProp}, which has type \texttt{Prop -$>$ Prop -$>$ Prop}.
  \item Linearize the resulting tree to obtain a string of the form ``A and B'', or a corresponding sentence in some other language.
  \end{enumerate}

\noindent
When iterated, these rules can produce ambiguous sentences.
For instance, the sentence
\begin{center}
\textit{for all numbers $ x $, $ x $ is even or $ x $ is odd}
\end{center}
is syntactically ambiguous between

\begin{verbatim}
  forall Num (x => or (even x) (odd x))
  \end{verbatim}
and

\begin{verbatim}
  or (forall Num (x => even x)) (odd x)
  \end{verbatim}
The latter interpretation can be filtered out by type checking, because it contains an unbound variable (unless that variable is bound from outside).
A normal writer would probably rely on this resolution of the ambiguity; if not, the grammar can also express the same proposition without syntactic ambiguity, by using a disjunction of adjectives:
\begin{center}
\textit{for all numbers $ x $, $ x $ is even or odd}
\end{center}

Another example of ambiguity is propositions containing both \textit{and} and \textit{or}. 
A CNL could stipulate, similarly to many programming languages, that \textit{and} binds stronger than \textit{or}.
Thus \textit{A and B or C} would always mean (\textit{A and B}) \textit{or C}, where \textit{A, B, C} are expressions of the same category, such as sentence, adjective, or noun phrases.
But it is not guaranteed that all readers know this rule.
Informath tries to avoid such stipulations and instead recognize potential ambiguities and find ways to resolve them.
One way to resolve the scope of conjunctions is \textbf{discontinuous conjunctions}, \textit{both - and} or \textit{either - or}.
They work like parentheses around the parts that belong together:

\begin{compactitem}
\item \textit{either A and B or C} means (\textit{A and B}) \textit{or C},
\item \textit{A and either B or C} means \textit{A and} (\textit{B or C)}.
\end{compactitem}
\subsection{Aggregation}

The disjunction of adjectives is an instance of \textbf{aggregation} in the NLG sense.
It can be defined by a rule that says that the common subject of two adjectives can be shared.
In GF, it is a conversion between abstract syntax trees, based on the functions

\begin{verbatim}
  fun OrProp : Prop -> Prop -> Prop
  fun OrAdj : Adj -> Adj -> Adj
  fun AdjProp : Adj -> Exp -> Prop
  \end{verbatim}
The aggregation rule can be expressed as follows with abstract syntax:

\begin{verbatim}
  OrProp (OrProp adj1 exp) (OrProp adj2 exp)
    ---> AdjProp (OrAdj adj1 adj2) exp
  \end{verbatim}
or, as we will usually write, in a quasi-context-free format

\begin{verbatim}
  <Exp> is <Adj>1 <Conj> <Exp> is <Adj>2
      --->   <Exp> is <Adj>1 <Conj> <Adj>2
  \end{verbatim}
The conjunction \texttt{$<$Conj$>$} can be \textit{and} or \textit{or}.

A related aggregation rule shares a common adjective of two subjects, as in \textit{$x$ and $y$ are prime}:

\begin{verbatim}
  <Exp>1 is <Adj> <Conj> <Exp>2 is <Adj>
      --->   <Exp>1 <Conj> <Exp>2 are <Adj>
  \end{verbatim}
Notice that the choice between ``is'' and ``are'' is not hard-coded but depends on grammatical agreement, which is determined by the subject and the conjunction together.
It is one of the things automatically guaranteed by the RGL.

The actual aggregation rules of Informath generalize the above rules in numerous ways:

\begin{compactitem}
\item there can be more than two subjects and adjectives,
\item verbs and nouns have similar rules,
\item an \texttt{$<$Adj$>$} can be formed by applying an \texttt{$<$Adj2$>$} to an argument, which creates a possibility for aggregation, as in \textit{$6$ is even and divisible by $3$}.
\end{compactitem}

\subsection{Negation}

A generic negation, applicable to any proposition, is the prefix \textit{it is not the case that $A$}.
If $A$ is formed by a conjunction, a scope ambiguity arises, as in
\begin{center}
\textit{it is not the case that $n$ is odd and $n$ is prime}
\end{center}
which can be of both the form $\sim \!(A \& B)$ and the form $(\sim \!A) \& B$.
Aggregation can, again, help express one alternative reading (the first one) unambiguously:
\begin{center}
\textit{it is not the case that $n$ is odd and prime}
\end{center}
The second reading is achieved by using a negation of the predicate:
\begin{center}
\textit{$n$ is not odd and $n$ is prime.}
\end{center}
In general, predicate negation is preferable whenever possible, also because it is more concise.
The rules that generate it for adjectives and verbs are

\begin{verbatim}
  it is not the case that <Exp> is <Adj>
      --->   <Exp> is not <Adj>
  it is not the case that <Exp> <Verb>
      --->   <Exp> does not <Verb>
  \end{verbatim}
An even more compact form of negations is \textbf{antonyms}, predicates defined as negations of each other.
They can be both verbal (\textit{even - odd}) and symbolic ($=$ -- $\neq$).
Antonym relations could be defined in symbol tables and checked to be consistent with the actual definitions in the formal code; this is currently a matter of future work.

\subsection{In situ quantification}

In addition to aggregation, a useful fluency mechanism is \textbf{in situ quantification}: quantifiers that appear as syntactic arguments (such as subjects and objects) of predicates (\textit{in situ} is Latin for ``in the place''):
\begin{center}
\textit{every natural number $ n $ is even or odd}
\end{center}
If the variable $n$ is not used elsewhere, as is the case here, it can be dropped:
\begin{center}
\textit{every natural number is even or odd}
\end{center}
The most important condition is that the variable must appear exactly once in the scope of the quantifier.
This is not the case with \textit{$ n $ is even or $ n $ is odd}, but it works in the aggregated version \textit{$ n $ is even or odd}.
Applying aggregation before in situ is a typical example of how one NLG operation can make another one possible.

In situ quantification in the presence of a negation is tricky.
Unrestricted NLG rules would produce all of the following:

\begin{enumerate}
\item \texttt{not (forall Num (n =$>$ even n))} \textit{every number is not even}
\item \texttt{forall Num (n =$>$ not (even n))} \textit{every number is not even}
\item \texttt{not (exists Num (n =$>$ even n))} \textit{some number is not even}
\item \texttt{exists Num (n =$>$ not (even n))} \textit{some number is not even}
\end{enumerate}

\noindent
When polling around, we have found that some readers consider (1) and (4) to be correct translations.
Thus \textit{every} and \textit{some} would have different behaviours with respect to negation.
But some others prefer (2) to (1).
Other quantifier words, \textit{all}, \textit{each}, \textit{a}, \textit{any}, and \textit{a certain}, may have different preferences \cite{hintikka-1979}, and individual readers may have their own interpretations.
The safest way to deal with this situation is not to generate such alternatives in the informalization mode, and to return ambiguous results in the formalization mode.

The baseline in informalization is thus to use explicit quantifier and negation prefixes.
However, if the quantifier is in the scope of the prefix negation, it can safely be made in situ:

\begin{compactitem}
\item \texttt{not (forall Num (n =$>$ even n))} \textit{it is not the case that every number is even}
\item \texttt{not (exists Num (n =$>$ even n))} \textit{it is not the case that some number is even}
\end{compactitem}
Likewise, if the predicate negation has an informalization by an antonym, it can be used with an in situ quantifier:

\begin{compactitem}
\item \texttt{forall Num (n =$>$ not (even n))} \textit{every number is odd}
\item \texttt{exists Num (n =$>$ not (even n))} \textit{some number is odd}
\end{compactitem}
Yet another device is the negative quantifier \textit{no}.
It can be interpreted as both ``for all not''  and ``not exists'':

\begin{compactitem}
\item \texttt{forall Num (n =$>$ not (even n))} \textit{no number is even}
\item \texttt{not (exists Num (n =$>$ even n))} \textit{no number is even}
\end{compactitem}
These two propositions are equivalent even constructively.
However, since they are intensionally different (have different proof objects), one might want to rule out expressions that are ambiguous between them.

To add even more complication, a sentence can contain more than one in situ quantifier:
\begin{center}
\textit{every first-degree equation has a real solution}
\end{center}
Such sentences do occur in mathematics, but their formalization can be tricky because of scope ambiguities.
The order is often left to right, but not always.
Ever since the pioneering work of Montague \cite{montague-1973}, in situ quantifier scopes have been a central topic in the literature of linguistic semantics, for instance \cite{cooper,hintikka-1979}.
Since there is no safe way to resolve ambiguities, Informath is cautious in generating them as a part of NLG.
But the Informath grammar does cover them, in the interest of linguistic completeness.
It can therefore parse them and leave their disambiguation to a separate phase by using a mechanism called Cooper storage \cite{cooper}.

Scope problems also appear with quantifier prefixes moved after the body of quantification.
If there is just one quantifier, as in
\begin{center}
\textit{the product of $n$ and $n + 1$ is even for all natural numbers $n$}
\end{center}
there is no problem.
Post-quantifier sometimes provide a good way to avoid ``subsequent dollars''.
But if the sentence also has a prefixed quantifier, the order of quantifiers is ambiguous: 
\begin{center}
\textit{for all numbers $m$, the product of $m$ and $n$ is even for some number $n$}
\end{center}
However, if the prefix is expressed in a separate sentence, no ambiguity arises:
\begin{center}
\textit{Let $m$ be a natural number. Then $m \times n$ is even for some number $n$.}
\end{center}
Again, the grammar must be able to recognize all variants, even the ambiguous ones, and let other components of the system resolve ambiguities and prevent the generation of them.

\subsection{Judgements}

The level above propositions is \textbf{judgements}, belonging to the category \texttt{Jmt} in the Informath grammar.
The natural verbalization of a judgement depends on the linguistic type of the defined constant --- whether its value type is \texttt{Exp}, \texttt{Kind}, \texttt{Prop}, or \texttt{Proof}.

Starting with \texttt{Exp}, we have, for example,

\begin{compactitem}
\item \texttt{def square : (x : Elem Num) -$>$ Elem Num := x =$>$ times x x.}
\item \textit{Definition. Let $x$ be a number. Then the square of $x$ is a number defined as $x x$.}
\end{compactitem}
If the definiens is missing, the part starting with \textit{defined} is dropped, so that just the type is expressed.

For \texttt{Kind}, we can use the conventional indefinite article, even though the effect of a definition is universal:

\begin{compactitem}
\item \texttt{def Nat : Set := suchthat Int (n =$>$ Geq n 0).}
\item \textit{Definition. A natural number is an integer $n$ such that $n \geq 0$.}
\end{compactitem}
If the definiens is missing, the basic way to express this is \textit{Natural numbers are a basic type.}

For \texttt{Prop}, we use the conventional \textit{if}, even though the effect is \textit{if and only if}:

\begin{compactitem}
\item \texttt{def divisible : (n : Elem Nat) -$>$ (m : Elem Nat) -$>$ Prop}
\texttt{:= n =$>$ m =$>$ exists Nat (k =$>$ Eq n (times k m)).}
\item \textit{Definition. Let $n$ and $m$ be natural numbers. Then $n$ is divisible by $m$, if $n = k m$ for some natural number $k$.}
\end{compactitem}
If the definiens is missing, the basic way to express this is \textit{Then we can say that $n$ is divisible by $m$.}

For \texttt{Proof}, the defined constant itself can be used as the label of the text:

\begin{compactitem}
\item \texttt{def thm1 : (n : Elem Nat) -$>$ Proof (even n) -$>$ Proof (odd (plus n 1)) := n =$>$ p =$>$ ax2 n p.}
\item \textit{Theorem 1. Let $n$ be an even natural number. Then $n + 1$ is odd. Proof. Assume $n$ is an even number ($p$). Then $n + 1$ is odd by Axiom 2 applied to $n$ and $p$.}
\end{compactitem}
If the definiens is missing (like for an axiom), the \textit{Proof} part is just omitted.

The general schema for translating Dedukti judgements assumes the following \textbf{normal form of judgements}:
$$
\text{def} \; c : (x_1 : T_1) \rightarrow \cdots \rightarrow (x_n : T_n) \rightarrow T :=
  x_1 \Rightarrow \cdots \Rightarrow x_n \rightarrow d \, a_1 \ldots a_m.
$$
where

\begin{compactitem}
\item the bound variables in the type are the same as in the defining abstraction,
\item the value type $T$ is a basic type.
\end{compactitem}
(Rewrite rules will be discussed separately.)
As the translation algorithm of Informath \textit{does} cover all Dedukti judgements, this normal form is not required.
But it has the virtue of making the translations more fluent and natural-sounding in the presence of lexical functions.
Conversion to this form is therefore applied as a preparatory step of translation, together with dropping and permuting arguments in accordance with the symbol table.

If a Dedukti judgement is in the normal form, we can distinguish the following parts:

\begin{compactitem}
\item the defined constant $c$
\item the hypotheses $x_1 : T_1, \ldots, x_n : T_n$
\item the target type $T$
\item the optional definiens $d \, a_1 \ldots a_m$
\end{compactitem}
The grammar has two kinds of hypotheses:

\begin{compactitem}
\item \texttt{Let $<$Ident$>$ be a $<$Kind$>$} 
\item \texttt{Assume that $<$Prop$>$ ($<$Ident$>$)?}
\end{compactitem}
The choice depends thus on whether the type in the hypothesis is ``set-like'' or ``proposition-like'' --- whether it corresponds to the Informath type \texttt{Kind} or \texttt{Prop}.
In Dedukti code, two kinds of hypotheses also occur: ones with a bound variable, as in \texttt{(x : A) -$>$ B} and ones without, as in \texttt{A -$>$ B}.
The choice usually depends on whether the subsequent expression depends on the variable.
Proposition-like hypotheses seldom have variables, because the type does not depend on them; if a variable occurs, it is verbalized as a label given in parentheses after the proposition.
Set-like hypotheses need variables more often, but are typically left out in Dedukti when the value type does not depend on them.
However, a natural-sounding verbalization of a definition usually does need variables, which are therefore created as a part of the normalization that starts the translation process.

Hypotheses can be aggregated, in two principal ways:

\begin{compactitem}
\item shared Kind, of several variables: \textit{let k, m and n be natural numbers}
\item variable followed by a property: \textit{let p be a prime number}
\end{compactitem}
The definition part may also consist of a set of rewrite rules. 
They are translated into a GF expression that linearizes into an itemized list of cases:

\begin{verbatim}
  def plus : Elem Nat -> Elem Nat -> Elem Nat.
  [m] plus m 0 --> m
  [m, n] plus m (Succ n) --> Succ (plus m n).
  \end{verbatim}
translates to

\begin{compactitem}
\item Let $x$ and $y$ be natural numbers. Then the sum of $x$ and $y$ is a natural number. By cases:
  \begin{compactitem}
  \item for $m$, the sum of $m$ and $0$ is $m$. 
  \item for $m$ and $n$, the sum of $m$ and the successor of $n$ is the successor of the sum of $m$ and $n$. 
  \end{compactitem}
\end{compactitem}
\subsection{Proofs}

\label{proofs}

The treatment of proofs is still largely work in progress in Informath.
In addition to proof by cases modelled by rewrite rules, Dedukti proofs are composed of abstractions and applications.
The following rules are, hence, as such complete for all proof objects that can be formed in Dedukti:

\begin{compactitem}
\item identifiers and their applications:
  $$(f \, p_1 \ldots p_n) : P \;\;\Longrightarrow\;\; p_1 \ldots p_n . \text{ By } f, P.$$
\item abstractions:
  $$B_1 \Rightarrow \ldots \Rightarrow B_n \Rightarrow (p : P) \;\;\Longrightarrow\;\;  B_1 \ldots B_n . p . \text{ We conclude } P.$$
\end{compactitem}
Some type annotations of normal Dedukti code may be needed to achieve this format: the conclusion $P$ in each case, as well as the types of each variable in the bindings $B_i$, to turn them into hypotheses of the same forms as are used in the informalizations of definitions.

The extraction of texts from proof objects follows an old idea \cite{coscoy-al-1995}, previously adapted to a GF-like setting in \cite{ranta-1996}.
In that work, raw abstractions and applications are extended by special rules for natural deduction and with aggregation of, for instance, several conjunction introductions in one step.
However, a more modular way to informalize proofs should work more on the content determination and text planning phrases of NLG (cf.\ Section~\ref{nlg}).

Proof verbalization is to some extent a separate concern from the linguistic renderings of individual proof steps.
Those renderings are taken care of by the grammar, but their ordering and omission pose a new set of problems.
In addition, proofs exhibit interesting discourse structures such as backward and forward references \cite{zinn-2004}.
Following the work that defined the original challenged addressed by this paper \cite{jiang-al-2024}, we have left out a further discussion of proofs and concentrated on propositional content.

\section{Some first results}

\label{results}

The first version of Informath  \cite{ranta-2024} consisted of a multilingual mathematics lexicon and a CNL grammar. 
The lexicon was extracted from mathematical terms and their labels in Wikidata \cite{vrandecic-kroetzsch-2014}, in particular those collected in MathGloss \cite{horowitz-depaiva-2023}.
This lexicon has 5300, mostly multiword, terms with linearizations to eight languages, but only covering parts of the lexicon in most languages because of gaps in Wikidata labels.
Its abstract syntax consists of the unique Wikidata identifiers.
The CNL syntax was based on ForTheL \cite{paskevich-2007}, Naproche \cite{cramer-al-2009}, and GF-Lean \cite{pathak-2024} and had concrete syntaxes to the same eight languages as the lexicon.
The grammar was capable of parsing and translating a large fragment of Naproche statements but left out the proofs.
It was not yet equipped with actual informalization, but it was partially translated to Lean via the pipeline in \cite{pathak-2024}.

The current system, focused on the informalization of Dedukti, has made the grammar more abstract and increased its coverage.
It is semantically complete with respect to Dedukti, but still far from linguistic completeness.
It is compatible with the multilingual lexicon extracted from Wikidata, but the need of storing multiword terms in a GF grammar has decreased due to productive rules for multiword items accessible from symbol tables.
Thus the most important lexicon is the one containing single words extracted from the multiword lexicon and other sources, currently covering 4000 abstract words.
Almost all of them have linearizations to English, whereas the other languages have a more limited coverage. 
Only four languages (English, French, German, and Swedish) are currently included, since the focus has been on the formal side rather than on coverage of informal languages.
In addition, three languages (Czech, Finnish, and Polish) have compilable but so far unchecked grammars produced in interaction with a Claude LLM.

The informalization of code generated from Agda, Lean, and Rocq to Dedukti by third-party software has been tested with Informath.
Though not yet a part of the Informath software, it should be straightforward to include Informath symbol table entries as comments in files written in those systems, to provide direct informalizations for code written in them.
The most substantial informalizations have been produced from code natively written in Dedukti, with focus on undergraduate mathematics including arithmetic, set theory, algebra, and topology.
A subset of the ``100 theorems'' of \cite{wiedijk-2025} is used as a standing demo and test set.

A larger case study is the informalization of a homotopy type theory book \cite{rijke-2025}, meant as an experiment on more advanced mathematics than before \cite{ohlsson-ranta-2026}.
This study revealed an interesting issue with the third-party Agda to Dedukti translator: the generated code has to make explicit a lot of infrastructure, such as universe levels and coercions between them, which seriously blows up the code.
This infrastructure is needed to make sure that everything that type checks in Agda is also valid in Dedukti.
But it makes the code so complex that the author chose to convert the Agda code to Dedukti manually.
After this, the linearization of statements and definitions was mainly smooth, except for some interesting uses of the propositions as types principle, which required symbol tables where one and the same constant is expressed both as a Prop and as a Kind.
An example is

\begin{verbatim}
  htpy : "X is homotopic to Y" | "homotopy between X and Y" 
  \end{verbatim}
The choice between these wordings posed new challenges to NLG and the ranking of variants, both in the choice of the best variants and in the sheer number of alternatives.

Informath has also been used to generate synthetic data for machine learning, which was the task addressed in the paper that posed our original challenge \cite{jiang-al-2024}.
The experiment reported in \cite{huang-al-2025} used Informath to fine-tune an LLM to improve the autoformalization to Agda.
It achieved clear improvements above the baseline (the bare LLM Qwen2.5-7B-Instruct).
While this use of informalization is an established technique since \cite{wang-al-2018}, its importance seems to be decreasing as new methods of autoformalization by coding agents are gaining ground \cite{urban-2026}.
In this workflow, as stated in Section~\ref{introduction}, informalization itself seems to be the most relevant task, which is to produce readable explanations of the results of autoformalization.

\section{Conclusions}

\label{conclusions}

The obvious thing to conclude with is to ask how well we are doing with the three goals mentioned in the title of this paper:

\begin{compactitem}
\item Fluency: rich vocabulary, mixtures of symbolic and verbal expressions, and compression techniques such as aggregation and in situ quantification are able to produce natural-looking definitions and statements beyond a basic CNL level.
But the full coverage of the ordinary language of mathematics is still a far-away goal.
The main open problems are in the ranking of growing numbers of alternatives and in the informalization of proofs.
\item Productivity: the use of GF and its resource grammar library RGL makes it feasible to write linguistically accurate grammars that cover a wide variety of linguistic structures, even in languages other then English.
Symbol tables with lexical annotations supported by a large lexicon provide a low-code way to apply Informath to new mathematical concepts.
More automation in the extraction of terminology of texts would still be desirable when dealing with the autoformalization of large texts as opposed to interactively developed formalization.
\item Multilinguality: the third-party conversions from other formalisms to Dedukti already make it possible to informalize Agda, Lean, and Rocq.
All of these systems are based on constructive type theory, which could make it natural to convert them to Dedukti.
However, their different additional features such as universe hierarchies, structure types, type classes, and built-in types often make the Dedukti conversions explode in size and complexity, which means that we have not yet managed to reach the same level of fluency as with native Dedukti code.
Systems not based on type theory, such as HOL Light, Isabelle, Megalodon, and Mizar pose new kinds of problems, which are currently being investigated.
Multilinguality on the natural language side seems to be much smoother, as can be expected from earlier projects using GF in multilingual CNL systems \cite{angelov-ranta-2009}, and recently encouraged by experiments using LLMs to generate GF code.
\end{compactitem}
What about the common belief that informalization is \textit{easier} than formalization?
The work with Informath has certainly shown that informalization has its own problems, some of which we had not even expected.
What is more, informalization is clearly \textit{more difficult} than formalization in one respect: the results of formalization can be checked automatically, whereas informalization cannot.
We take this to imply that informalization should be approached with tools that are inherently reliable and explainable, and which inevitably require some manual work and expert knowledge.

\section{Future work}

\label{futurework}

Informath is --- and will continue to be in the foreseeable future --- work in progress.
Hence we want to mention just a couple of items for future work that seem important and are not so obvious that they can be covered by short mentions, such as extending the grammar, adding formalisms, adding natural languages, performing more extensive case studies, or integrating with different proof systems.

\textbf{Automation}: a controlled use of LLMs is a promising source of automating grammar building.
Initial experiments show that it is possible to extend the autoformalization workflow of \cite{urban-2026} so it constructs, in addition to the machine-checkable proof, a symbol table enabling the informalization of the formal proof.
This symbol table must of course be checked independently to guarantee reliability.
But checking a limited set of lexical items --- or any symbolic rules --- is work that can save us from checking unlimited amounts of run-time LLM output.

\textbf{NLG ranking}: the number of variants generated in the NLG pipeline is a problem that gets worse as the grammar is extended.
Since ranking is based on sorting by penalty, its complexity is worse than linear.
A similar explosion happens when Informath is used for parsing natural language.
For the parsing problem, a general solution exists for GF, pruning alternatives early based on statistical ranking \cite{angelov-ljunglof-2014}.
But it would be desirable to include dependent type checking in the pruning set-up.
Developing a similar algorithm to the NLG pipeline would be desirable and also possible, since it is also a task where a large set of trees is built incrementally.

\textbf{Proofs}: as mentioned in Section~\ref{proofs}, improving the informalization of proofs is a high priority task.
A promising model for this is the conversion from proof trees (in the sense of Gentzen) to linear proofs (as developed by Ja\'skowski, Fitch, and Prawitz) in natural deduction \cite{vonplato-2017}.
The ordering of the proof steps can then be studied separately, as well as the omission of steps that are ``obvious'' \cite{davis-1981}.
The ultimate challenge of proof informalization is to invert the \textbf{de Bruijn factor}: since formalization blows up the proof size by a factor of 10 to 20 \cite{debruijn-1980}, informalization should drop at least 90\% of the formal proof.

\subsection*{Acknowledgements}

I am grateful to many many people.
The first to mention are the contributors to Informath software: Adrian De Lon, Amélie Ledein, Inari Listenmaa, Jan Frederik Schaeffer, May Ohlsson, Shashank Pathak, and Soline du Crest.
The second group, much larger, are all the people who have shared their knowledge and resources: Chad Brown, Frédéric Blanqui, Hugo Herbelin, Jan von Plato, Josef Urban, Lucy Horowitz, Mario Carneiro, Michael Kohlhase, Paul-André Melliès, Per Martin-Löf, Peter Dybjer, Peter Koepke, Philippe de Groote, Rishikesh Vaishnav, Thierry Coquand, and Valeria de Paiva.
This work has been supported by the European Research Council through the grants \textit{NextReason} (Grant ID 101200949) and \textit{MALINCA} (Grant ID 101167526) as well as \textit{EuroProofNet} (COST action CA20111).


\bibliographystyle{plain}
\bibliography{gf-bib}

\end{document}